%% file: root.tex
\documentclass[letterpaper, 10 pt, conference]{ieeeconf}  

\IEEEoverridecommandlockouts                              

\title{\LARGE \bf Efficient Bimanual Manipulation Using Learned Task Schemas}
\author{
  \textbf{Rohan Chitnis$^*$ \hspace{12mm} Shubham Tulsiani \hspace{12mm} Saurabh Gupta \hspace{12mm} Abhinav Gupta}\\\\
  \small{MIT Computer Science and Artificial Intelligence Laboratory, Facebook Artificial Intelligence Research}\\
  \texttt{\small{ronuchit@mit.edu, shubhtuls@fb.com, saurabhg@illinois.edu, gabhinav@fb.com}}
  \thanks{\hspace*{-1em} * Work done during an internship at Facebook AI Research.}
}

\input{math_commands.tex}

\input{preamble.tex}

\input{variables.tex}

\begin{document}
\bibliographystyle{IEEEtran}
\maketitle
\thispagestyle{empty}
\pagestyle{empty}
\setlength{\textfloatsep}{8pt}

\begin{abstract}
  We address the problem of effectively composing skills to solve
  sparse-reward tasks in the real world. Given a set of parameterized
  skills (such as exerting a force or doing a top grasp at a
  location), our goal is to learn policies that invoke these skills to
  efficiently solve such tasks. Our insight is that for many tasks,
  the learning process can be decomposed into learning a
  state-independent task schema (a sequence of skills to execute) and
  a policy to choose the parameterizations of the skills in a
  state-dependent manner. For such tasks, we show that explicitly
  modeling the schema's state-independence can yield significant
  improvements in sample efficiency for model-free reinforcement
  learning algorithms. Furthermore, these schemas can be transferred
  to solve related tasks, by simply re-learning the parameterizations
  with which the skills are invoked. We find that doing so enables
  learning to solve sparse-reward tasks on real-world robotic systems
  very efficiently. We validate our approach experimentally over a
  suite of robotic bimanual manipulation tasks, both in simulation and
  on real hardware. See videos at \texttt{http://tinyurl.com/chitnis-schema}.
\end{abstract}

\input{introduction.tex}
\input{related-work.tex}
\input{approach.tex}
\input{experiments.tex}
\input{futurework.tex}



\section*{Acknowledgments}
We thank Dhiraj Gandhi for help with the experimental
setup. Rohan is supported by an NSF Graduate Research Fellowship through his capacity with the Massachusetts Institute of Technology. Any
opinions, findings, and conclusions expressed in this material are
those of the authors and do not necessarily reflect the views of our
sponsors.

\bibliography{references}

\end{document}

%% file: math_commands.tex
\usepackage{amsmath,amsfonts,bm}

\def\1{\bm{1}}

\DeclareMathAlphabet{\mathsfit}{\encodingdefault}{\sfdefault}{m}{sl}
\SetMathAlphabet{\mathsfit}{bold}{\encodingdefault}{\sfdefault}{bx}{n}



%% file: preamble.tex
\usepackage{multicol}
\usepackage{color}
\usepackage{amssymb}
\usepackage{amsmath}
\usepackage{tikz}
\usepackage{bbm}
\usetikzlibrary{arrows,decorations.markings}
\usepackage[makeroom]{cancel}

\usepackage{booktabs}
\usepackage{graphicx}
\usepackage{todonotes}

\usepackage[font=small]{caption}
\usepackage{subcaption}

\usepackage[noline]{algorithm2e}
\makeatletter
\newcommand{\removelatexerror}{\let\@latex@error\@gobble}
\makeatother

\usepackage{indentfirst}
\usepackage{float}
\usepackage{mathtools}

\newenvironment{tightlist}%
{\begin{list}{$\bullet$}{%
    \setlength{\topsep}{0in}
    \setlength{\partopsep}{0in}
    \setlength{\itemsep}{0in}
    \setlength{\parsep}{0in}
    \setlength{\leftmargin}{1.5em}
    \setlength{\rightmargin}{0in}
    \setlength{\itemindent}{-.1in}
}
}%
{\end{list}
}

\newcommand{\secref}[1]{Section~\ref{#1}}

\newcommand{\figref}[1]{Figure~\ref{#1}}
\newcommand{\algref}[1]{Algorithm~\ref{#1}}

\newcommand{\tabref}[1]{Table~\ref{#1}}

\DeclarePairedDelimiterX{\infdivx}[2]{(}{)}{%
  #1\;\delimsize\|\;#2%
}

\newcommand{\specialcell}[2][c]{%
  \begin{tabular}[#1]{@{}l@{}}#2\end{tabular}}

%% file: variables.tex
\newcommand{\A}{\mathcal{A}}

\newcommand{\X}{\mathcal{X}}

\newcommand{\pamdp}{{\sc pamdp}}
\newcommand{\mdp}{{\sc mdp}}
\newcommand{\ppo}{{\sc ppo}}

\newcommand{\M}{\mathcal{M}}

\newcommand{\data}{\mathcal{D}}

\newcommand{\Ex}{\mathbb{E}}

%% file: introduction.tex
\section{Introduction}
\label{sec:introduction}
Let us consider the task of opening a bottle. How should a two-armed
robot accomplish this? Even without knowing the bottle geometry, its
position, or its orientation, one can answer that the task will
involve holding the bottle's base with one hand, grasping the bottle's
cap with the other hand, and twisting the cap off. This ``schema,''
the high-level plan of what steps need to be executed, only depends on
the task and not on the object's geometric and spatial state, which
only influence how to parameterize each of these steps (e.g., deciding
where to grasp, or how much to twist).

\begin{figure}[t]
  \centering
    \noindent
    \includegraphics[width=\columnwidth]{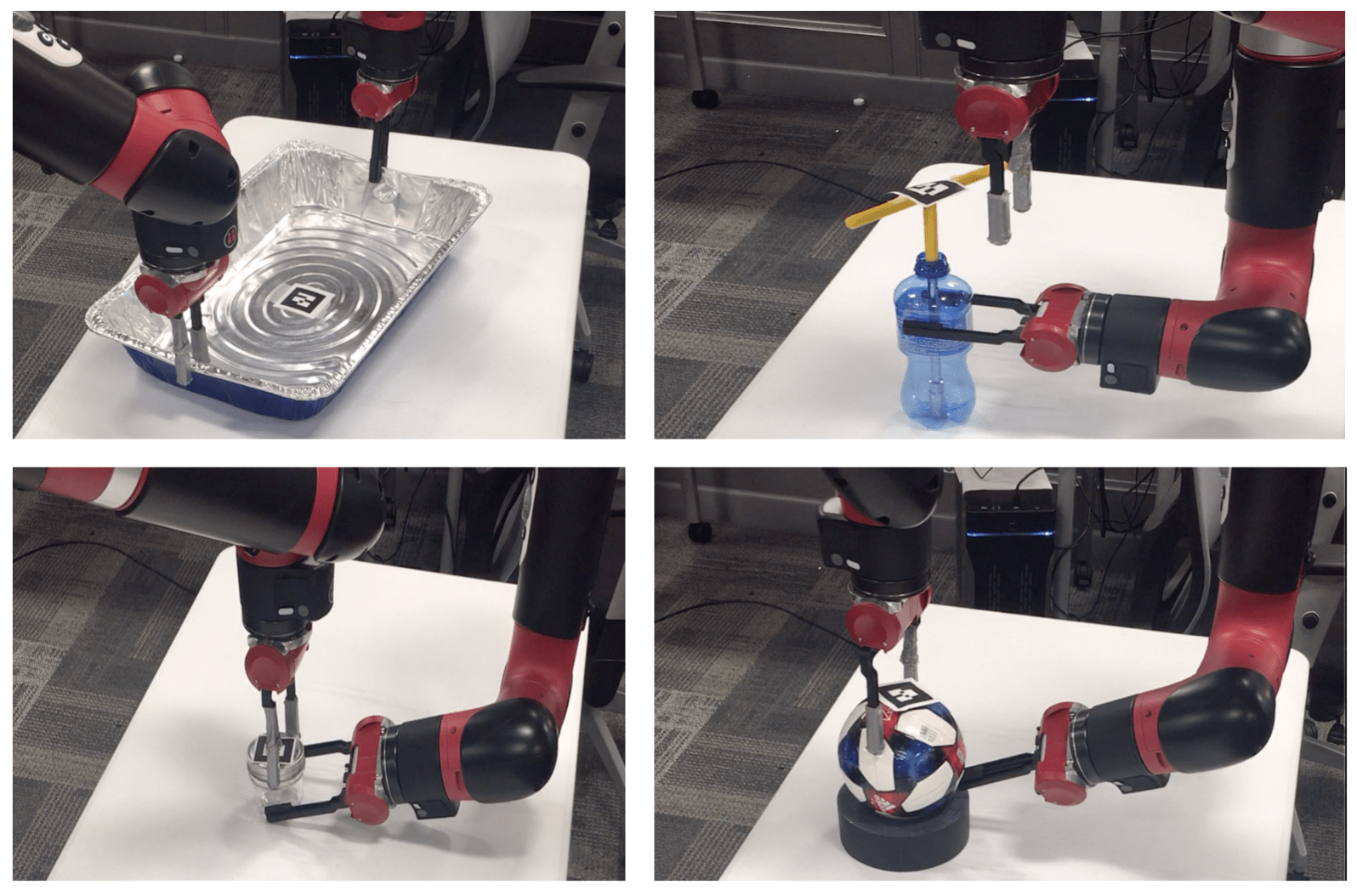}
    \caption{We learn policies for solving real-world sparse-reward
      bimanual manipulation tasks from raw RGB image observations. We
      decompose the problem into learning a state-independent task
      schema (a sequence of skills to execute) and a state-dependent
      policy that appropriately instantiates the skills in the context
      of the environment. This decomposition speeds up learning and
      enables transferring schemas from simulation to the real world,
      leading to successful task execution within a few hours of
      real-world training. The AR tags are only used for automating
      the reward computation; our model is not given any object pose
      information. \emph{Top left}: Lifting an aluminum
      tray. \emph{Top right}: Rotating a T-wrench. \emph{Bottom left}:
      Opening a glass jar. \emph{Bottom right}: Picking up a large
      soccer ball.}
  \label{fig:teaser}
\end{figure}

Reinforcement learning methods provide a promising approach for learning such behaviors from raw sensory input~\cite{dqn,visuomotor,ddpg}.
However, typical end-to-end reinforcement learning methods do not
leverage the schematics of tasks, and instead aim to solve tasks by
learning a policy, which would involve inferring both the schema and
the parameterizations, as a function of the raw sensory input. These
approaches have led to impressive successes across domains such as
game-playing~\cite{dqn,doubledqn,duelingdqn,a3c} and robotic control
tasks~\cite{visuomotor,naf,mbmf,trpo}, but are known to have
very high sample complexity. For instance, they require millions of
frames of interaction to learn to play Atari games, or several weeks'
worth of experience to learn simulated control policies, which makes
them impractical to train on real hardware.

In this work, we address the problem of learning to perform tasks in
environments with a sparse reward signal, given a discrete set of
generic \emph{skills} parameterized by continuous arguments. Examples
of skills include exerting a force at a location or moving an end
effector to a target pose.  Thus, the action space is hybrid
discrete-continuous~\cite{pamdp}: at each timestep, the agent must decide both 1)
which skill to use and 2) what continuous arguments to use
(e.g., the location to apply force, the amount of force, or the target
pose to move to). The sample inefficiency of current reinforcement
learning methods is exacerbated in domains with these large search
spaces; even basic tasks such as opening a bottle with two arms are
challenging to learn from sparse rewards. While one could
hand-engineer dense rewards, this is undesirable as it does not scale
to more complicated tasks.  We ask a fundamental question: can we use
the given skills to efficiently learn policies for tasks with a large
policy search space, like bimanual manipulation, given only sparse
rewards?

Our insight is that for many tasks, the learning process can be
decomposed into learning a \textit{state-independent} task schema
(sequence of skills) and a \textit{state-dependent} policy that
chooses appropriate parameterizations for the different skills. Such a
decomposition of the policy into state-dependent and state-independent
parts simplifies the credit assignment problem and leads to more
effective sharing of experience, as data from different instantiations of
the task can be used to improve the same shared skills. This leads to
faster learning.

This modularization can further allow us to \emph{transfer} learned
schemas among related tasks, \textit{even if they have different state
  spaces}. For example, suppose we have learned a good schema for
picking up a long bar in simulation, where we have access to object
poses, geometry information, etc. We can then reuse that schema for a
related task such as picking up a tray in the real world from only raw
camera observations, even though both the state space and the optimal
parameterizations (e.g., grasp poses) differ significantly. As the
schema is fixed, policy learning for this tray pickup task will be
very efficient, since it only requires learning the
(observation-dependent) arguments for each skill. Transferring the
schema in this way enables learning to solve sparse-reward tasks very
efficiently, making it feasible to train real robots to perform
complex skills. See \figref{fig:overview} for an overview of our
approach.

We validate our approach over a suite of robotic bimanual manipulation
tasks, both in simulation and on real hardware. We give the robots a
very generic library of skills such as twisting, lifting, and
reaching. Even given these skills, bimanual manipulation is
challenging due to the large search space for policy optimization.
We consider four task families: lateral lifting, picking, opening, and
rotating, all with varying objects, geometries, and initial poses. All tasks
have a sparse binary reward signal: 1 if the task is completed, and 0
otherwise. We empirically show that a) explicitly modeling schema
state-independence yields large improvements in learning efficiency
over the typical strategy of conditioning the policy on the full
state, and b) transferring learned schemas to real-world tasks allows
complex manipulation skills to be discovered within only a few hours
(\textless 10) of training on a single setup. \figref{fig:teaser}
shows some examples of real-world tasks solved by our system.

%% file: related-work.tex
\section{Related Work}
\label{sec:relatedwork}
\vspace{1mm}
\noindent \textbf{Search in parameterized action spaces.} An agent
equipped with a set of skills parameterized by continuous arguments
must learn a policy that decides both which skills to use and what
continuous arguments to use for them. Parameterized action \mdp s
(\pamdp s)~\cite{pamdp} were constructed for this exact problem
setting. Recent work has addressed deep reinforcement learning for
\pamdp s~\cite{pamdprl1,pamdprl2}, by learning policies that output
both the discrete skill and continuous parameter selections at each
timestep. In contrast, we propose a model that bakes in
state-independence of the discrete skill selection, and show that this
assumption not only improves learning efficiency, but also is
experimentally useful. A separate line of work learns control
policies for steps in a \emph{policy sketch}~\cite{policysketch},
which can be recombined in novel ways to solve new task instances;
however, this work does not consider the discrete search aspect of the
problem, as we do.


\begin{figure}[t]
  \centering
  \noindent
  \vspace{0.7em}
    \includegraphics[width=\columnwidth]{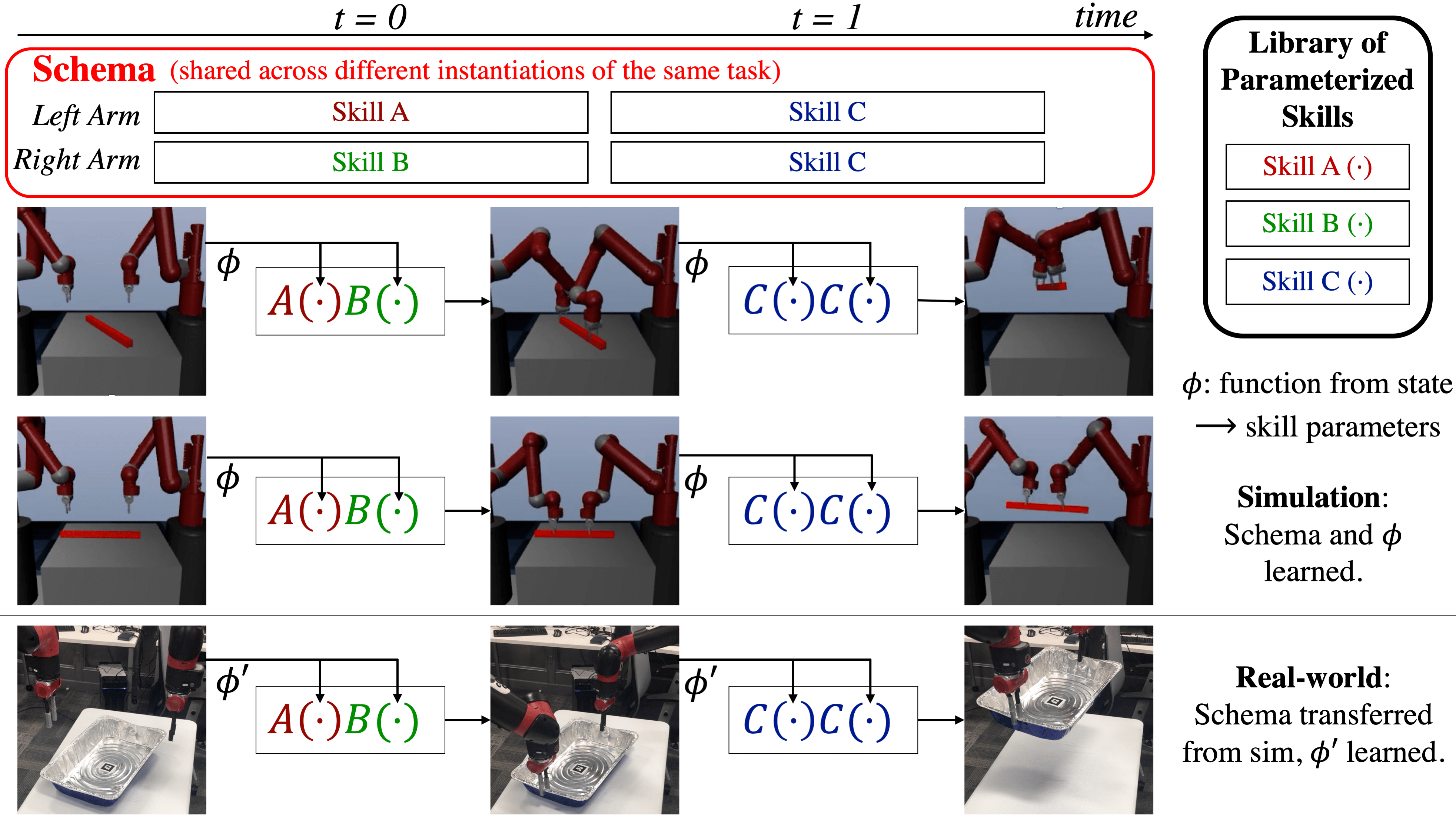}
    \caption{In bimanual manipulation tasks, a schema is a sequence of
      skills for each arm to execute. We train in simulation both a
      state-independent model for predicting this schema and a
      state-dependent neural network $\phi$ for predicting its
      continuous parameters, via reinforcement learning. Our
      experiments show that using a state-independent schema predictor
      for these tasks makes training significantly more efficient. To
      solve real-world tasks, we transfer schemas learned in
      simulation, and only optimize $\phi$.}
  \label{fig:overview}
\end{figure}

\vspace{1mm}
\noindent \textbf{Transfer learning for robotics.} The idea of
transferring a learned policy from simulation to the real world for
more efficient robotic learning was first developed in the early
1990s~\cite{transferold1,transferold2}. More recent techniques include
learning from model ensembles~\cite{transferensembles} and utilizing
domain
randomization~\cite{transferdomainrand1,transferdomainrand2,transferdomainrand3},
in which physical properties of a simulated environment are randomized
to allow learned policies to be robust.  However, as these methods
directly transfer the policy learned in simulation, they rely on the
simulation being visually and physically similar to the real world.
In contrast, we only transfer one part of our learned policy --- the
skill sequence to be executed --- from simulation to the real world,
and allow the associated continuous parameters to be learned in the
real-world domain.

\vspace{1mm}
\noindent \textbf{Temporal abstraction for reinforcement learning.}
The idea of using temporally extended actions to reduce the sample
complexity of reinforcement learning algorithms has been studied for
decades~\cite{temporal1,temporal2,temporal3,temporal4}. For instance,
work on macro-actions for \mdp s~\cite{temporal4} attempts to build a
hierarchical model in which the primitive actions occupy the lowest
level, and subsequently higher levels build local policies, each
equipped with their own termination conditions, that make use of
actions at the level below. More recent work seeks to learn these
hierarchies~\cite{frans2017meta,nachum2018data}, but successes have
largely been limited to simulated domains due to the large amount of
data required. In our work, we propose a model that makes skill
selection independent of state, enabling real-world robotic tasks to be solved via transfer.


%% file: approach.tex
\section{Approach}
\label{sec:approach}
Given a set of parameterized skills, we aim to solve sparse-reward
tasks by learning a policy that decides both which skill to execute
and what arguments to use when invoking it. Our insight is that, for
many tasks, the same sequence of skills (possibly with different
arguments) can be used to optimally solve different
instantiations of the task. We operationalize this by disentangling
the policy into a state-independent task schema (sequence of skills)
and a state-dependent prediction of how to parameterize these skills.
We first formally define our problem setup, and then present our model
for leveraging the state-independence of schemas to learn
efficiently. Finally, we describe how our approach also allows
transferring schemas across tasks, letting us learn real-world
policies from raw images by reusing schemas learned for related tasks
in simulation.

\input{exp-schemas.tex}

\vspace{1mm}
\noindent \textbf{Problem Setup.}  Each task we consider is defined as
a parameterized action Markov decision process
(\pamdp)~\cite{pamdp,mdp} with finite horizon $T$. The reward for
each task is a binary function indicating whether the current state is
an element of the set of desired goal configurations, such as a state
with the bottle opened. The learning objective, therefore, is to
obtain a policy $\pi$ that maximizes the expected proportion of times
that following it achieves the goal. Note that this is a particularly
challenging setup for reinforcement learning algorithms due to the
sparsity of the reward function.

The agent is given a discrete library of generic \emph{skills} $\X$,
where each skill $x \in \X$ is parameterized by a corresponding vector
$v^x$ of continuous values. Examples of skills can include exerting a
force at a location, moving an end effector to a target pose, or
rotating an end effector about an axis. Let $\A$ denote the
action space of the \pamdp. An action $a \in \A$ is
a tuple $\langle x, v^x \rangle$, indicating what skill to
apply as well as the corresponding parameterization. A \emph{schema}
$\bar{x}$ is a sequence of $T$ skills in $\X$, where
$\bar{x} = x_1, x_2, ..., x_T$ captures the sequence of skills but not
their continuous parameters.

\emph{Assumption.} We assume that the optimal schema $\bar{x}^*$ is
state-independent: it depends only on the task, not on the state and
its dynamics. This implies that the same schema is optimal for all
instantiations of a task, e.g. different geometries and poses of
objects. We note that this is a valid assumption across many tasks of
interest, since the skills themselves can be appropriately chosen to
be complicated and expressive, such as stochastic, closed-loop control
policies for guiding an end effector.

\vspace{1mm}
\noindent \textbf{Modular Policies.}  The agent must learn a policy
$\pi$ that, at each timestep, infers both which skill $x \in \X$ to
use (a discrete choice) and what continuous arguments $v^x$ to use.

\begin{figure}[t]
  \centering
  \noindent
  \vspace{0.7em}
    \includegraphics[width=\columnwidth]{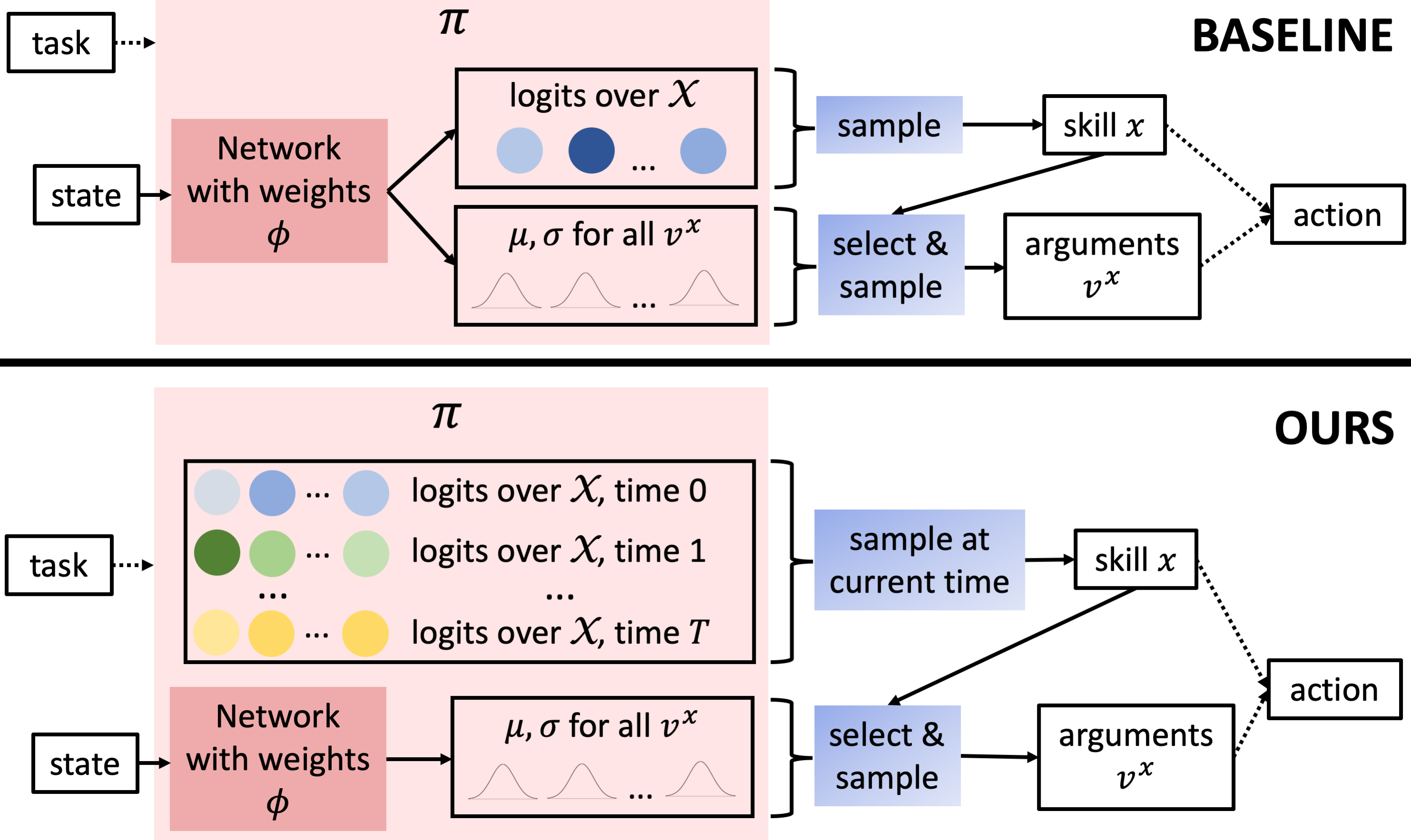}
    \caption{\emph{Top:} A baseline policy architecture for solving
      tasks with a discrete set of skills $\X$, each parameterized by
      a vector $v^x$ of continuous values. The policy for a task is a
      neural network with weights $\phi$ that predicts 1) logits over
      which skill to use and 2) mean and variance of a Gaussian over
      continuous argument values for all skills. To sample an action,
      we first sample a skill based on the logits, then select the
      corresponding subset of continuous arguments, and finally sample
      argument values from those Gaussians. \emph{Bottom:} Our
      proposed policy architecture, which leverages the assumption
      that the optimal schema is state-independent. The key difference
      is that the neural network only predicts a distribution over
      continuous argument values, and we train a state-independent
      $T \times |\X|$ array of logits over which skill to use at each
      timestep. In the text, we discuss how to update these
      logits.}
  \label{fig:model}
\end{figure}

What is a good form for such a policy? A simple strategy, which we use
as a baseline and depict in \figref{fig:model} (top), would be to
represent $\pi$ via a neural network, with weights $\phi$, that takes
the state as input and has a two-headed output. One head predicts
logits that represent a categorical distribution over the skills $\X$,
while the other head predicts a mean and variance of a Gaussian distribution over
continuous argument values for all skills. To sample
an action, we can sample $x \in \X$ from the logits predicted by the
first head, then sample arguments using the subset of means and
variances predicted by the second head that correspond to $v^x$.

However, this does not model the fact that the optimal schema is
state-independent. To capture this, we need to remove the dependence
of the discrete skill selection on the input state. Thus, we propose
to maintain a separate $T \times |\X|$ array, where row $t$ is the
logits of a categorical distribution over which skill to use at time
$t$. Note that $T$ is the horizon of the \mdp. In this architecture,
the neural network is only tasked with predicting the skill
arguments. The $T \times |\X|$ array of logits and the neural network,
taken together, represent the policy $\pi$, as depicted in
\figref{fig:model} (bottom).

\vspace{1mm}
\noindent \textbf{Learning Schemas and Skill Arguments.}  The weights
$\phi$ of the neural network can be updated via standard policy
gradient methods. Let $\tau$ denote a trajectory induced by following
$\pi$ in an episode. The objective we wish to maximize is
$J(\phi) \equiv \Ex_\tau[r(\tau)]$. Policy gradient methods such as
{\sc reinforce}~\cite{reinforce} leverage the likelihood ratio trick,
which says that
$\nabla_{\phi} J(\phi) = \Ex_\tau[r(\tau) \nabla_{\phi} \log
\pi(\tau)]$, to tune $\phi$ via gradient ascent. When estimating this
gradient, we treat the current setting of the array of logits as a
constant.

Updating the logits within the $T \times |\X|$ array can also be
achieved via policy gradients; however, since there is no input, and
because we have sparse rewards, the policy optimization procedure is
quite simple. Let $\varphi_{tx}$ be the logit for time $t$ and skill
$x$. Given trajectory
$\tau = \langle s_0, x_0, v_0^{x_0}, s_1, x_1, v_1^{x_1}, ..., s_T
\rangle$:
\begin{tightlist}
\item If $\tau$ achieves the goal, i.e. $r(\tau)>0$, increase
  $\varphi_{tx}$ for each timestep $t$ and skill $x$ taken at that
  timestep.
\item If $\tau$ does not achieve the goal, i.e. $r(\tau)=0$, decrease
  $\varphi_{tx}$ for each timestep $t$ and skill $x$ taken at that
  timestep.
\end{tightlist}

The amount by which to increase or decrease $\varphi_{tx}$ is absorbed
by the step size and thus gets tuned as a hyper-parameter. See
\algref{alg:learn} for full pseudocode.

\begin{algorithm}[t]
  \SetAlgoLined
  \SetAlgoNoEnd
  \DontPrintSemicolon
  \SetKwFunction{algo}{algo}\SetKwFunction{proc}{proc}
  \SetKwProg{myalg}{Algorithm}{}{}
  \SetKwProg{myproc}{Subroutine}{}{}
  \SetKw{Continue}{continue}
  \SetKw{Break}{break}
  \SetKw{Return}{return}
  \myalg{\textsc{Train-Policy}$(\M, \alpha, \beta)$}{
    \nl \textbf{Input:} $\M$, an \mdp\ as defined in \secref{sec:approach}.\;
    \nl \textbf{Input:} $\alpha$ and $\beta$, step sizes.\;
    \nl Initialize neural network weights $\phi$.\;
    \nl Zero-initialize $T \times |\X|$ array of logits $\varphi_{tx}$.\;
    \nl \While{not done}{
    \nl $\data \gets$ batch of trajectories $\tau=\langle s_t, x_t, v_t^{x_t} \rangle$ obtained from running policy $\pi$ in $\M$.\;
    \nl $\nabla_{\phi} J(\phi) \gets$ \textsc{PolicyGradient}$(\pi, \data)$\;
    \nl $\phi \gets \phi+\alpha \nabla_{\phi} J(\phi)$ \tcp*{\footnotesize Or Adam~\cite{adam}.}
    \nl \For{each trajectory $\tau \in \data$}{
    \nl \For{each skill $x_t$ used in $\tau$}{
    \nl \eIf{$\tau$ achieves the goal}{
    \nl $\varphi_{tx} \gets \varphi_{tx}+\alpha$
    }{
    \nl $\varphi_{tx} \gets \varphi_{tx}-\beta$
    }}}}}\;
\caption{Training policies $\pi$ that explicitly model
  the state-independence of schemas via a $T \times |\X|$ array of
  logits over what skill to use at each timestep.}
\label{alg:learn}
\end{algorithm}

\vspace{1mm}
\noindent \textbf{Schema Transfer Across Tasks.}  Since we have
disentangled the learning of the schema from the learning of the skill
arguments within our policy architecture, we can now \emph{transfer}
the $T \times |\X|$ array of logits across related tasks, as long as
the skill spaces and horizons are equal. Therefore, learning for a new
task can be made efficient by reusing a previously learned schema,
since we would only need to train the neural network weights $\phi$ to
infer skill arguments for that new task.

Importantly, transferring the schema is reasonable even when the tasks
have \emph{different state spaces}. For instance, one task can be a
set of simulated bimanual bottle-opening problems in a low-dimensional
state space, while the other involves learning to open bottles in the
real world from high-dimensional camera observations. As the state
spaces can be different, it follows immediately that the tasks can
also have different optimal arguments for the skills.

%% file: exp-schemas.tex
\begin{table*}
  \vspace{0.7em}
  \centering
  \resizebox{1.0\linewidth}{!}{
  \tabcolsep=0.08cm{
  \begin{tabular}{llll}
    \toprule[1.5pt]
    \textbf{Task Family} & \textbf{Object (Sim)} & \textbf{Objects (Real)} & \textbf{Schema Discovered from Learning in Simulation} \\
    \midrule[2pt]
    lateral lifting & bar & aluminum tray, rolling pin, heavy bar, plastic box & \textbf{1)} L: top grasp, R: top grasp \textbf{2)} L: lift, R: lift\\
    \midrule
    picking & ball & soccer ball & \textbf{1)} L: top grasp, R: go-to pose \textbf{2)} L: no-op, R: go-to pose \textbf{3)} L: lift, R: lift\\
    \midrule
    opening & bottle & glass jar, water bottle & \textbf{1)} L: top grasp, R: side grasp \textbf{2)} L: twist, R: no-op\\
    \midrule
    rotating & corkscrew & T-wrench, corkscrew & \textbf{1)} L: go-to pose, R: side grasp \textbf{2)} L: go-to pose, R: no-op \textbf{3)} L: rotate, R: no-op\\
    \bottomrule[1.5pt]
  \end{tabular}}}
  \caption{Task families, object considered in simulation, objects considered in real world, and schemas (for left and right arms) discovered by our algorithm in simulation. Schemas learned in simulation for a task family are transferred to multiple objects in the real world.}
  \label{tab:schemas}
\end{table*}

%% file: experiments.tex
\section{Experiments}
\label{sec:experiments}
We test our proposed approach on four robotic bimanual manipulation
task families: lateral lifting, picking, opening, and
rotating. \tabref{tab:schemas} lists the different objects that we
considered for each one. These task families were chosen because they
represent a challenging hybrid discrete-continuous search space for
policy optimization, while meeting our requirement that the optimal
schema is independent of the state. We show results on these tasks
both in simulation and on real Sawyer arms: schemas are learned in
simulation by training with low-dimensional state inputs, then
transferred as-is to visual inputs (in simulation as well as in the
real world), for which we only need to learn skill arguments. Our
experiments show that our proposed approach is significantly more
sample-efficient than one that uses the baseline policy architecture,
and allows us to learn bimanual policies on real robots in less than
10 hours of training. We first describe the experimental setup, then
discuss our results.

\subsection{MuJoCo Experimental Setup}
\vspace{1mm}
\noindent \textbf{Environment.} For all four task families, two Sawyer
robot arms with parallel-jaw grippers are placed at opposing ends of a
table, facing each other. A single object is placed on the table, and
the goal is to manipulate the object's pose in a task-specific way.
\emph{Lateral lifting (bar):} The goal is to lift a heavy and long bar
by 25cm while maintaining its orientation. We vary the bar's location
and density.  \emph{Picking (ball):} The goal is to lift a slippery
(low coefficient of friction) ball vertically by 25cm. The ball slips
out of the gripper when grasped by a single arm. We vary the ball's
location and coefficient of friction.  \emph{Opening (bottle):} The
goal is to open a bottle implemented as two links (a base and a cap)
connected by a hinge joint. If the cap is twisted without the base
being held in place, the entire bottle twists. The cap must undergo a
quarter-rotation while the base maintains its pose. We vary the bottle's
location and size.  \emph{Rotating (corkscrew):} The goal is to rotate
a corkscrew implemented as two links (a base and a handle) connected
by a hinge joint, like the bottle. The handle must undergo a
half-rotation while the base maintains its pose. We vary the
corkscrew's location and size.

\vspace{1mm}
\noindent \textbf{Skills.} The skills we use are detailed in
\tabref{tab:skills}, and the search spaces for the skill parameters
are detailed in \tabref{tab:parameters}. Note that because we have two
arms, we actually need to search over a cross product of this space
with itself.


\vspace{1mm}
\noindent \textbf{State and Policy Representation.} Experiments conducted
in the MuJoCo simulator~\cite{mujoco} use a low-dimensional state:
proprioceptive features (joint positions, joint velocities, end
effector pose) for each arm, the current timestep, geometry
information for the object, and the object pose in the world frame
and each end effector's frame. The policy is represented as a 4-layer
MLP with 64 neurons in each layer, ReLU activations, and a
multi-headed output for the actor and the critic. Since object
geometry and pose can only be computed within the simulator, our
real-world experiments will instead use raw RGB camera images.

\input{exp-skills.tex}

\input{exp-skills-params.tex}

\vspace{1mm}
\noindent \textbf{Training Details.} We use the Stable
Baselines~\cite{stable-baselines} implementation of proximal policy
optimization (\ppo)~\cite{ppo}, though our method is agnostic to the
choice of policy gradient algorithm. We use the following
hyper-parameters: Adam~\cite{adam} with learning rate $0.001$,
clipping parameter $0.2$, entropy loss coefficient $0.01$, value
function loss coefficient $0.5$, gradient clip threshold $0.5$, number
of steps $10$, number of minibatches per update $4$, and number of
optimization epochs $4$.  Our implementation builds on the Surreal
Robotics Suite~\cite{robosuite}. Training is parallelized across 50
workers. The time horizon $T=3$ in all tasks.

\begin{figure}
  \vspace{0.7em}
\begin{center}
\includegraphics[width=\linewidth]{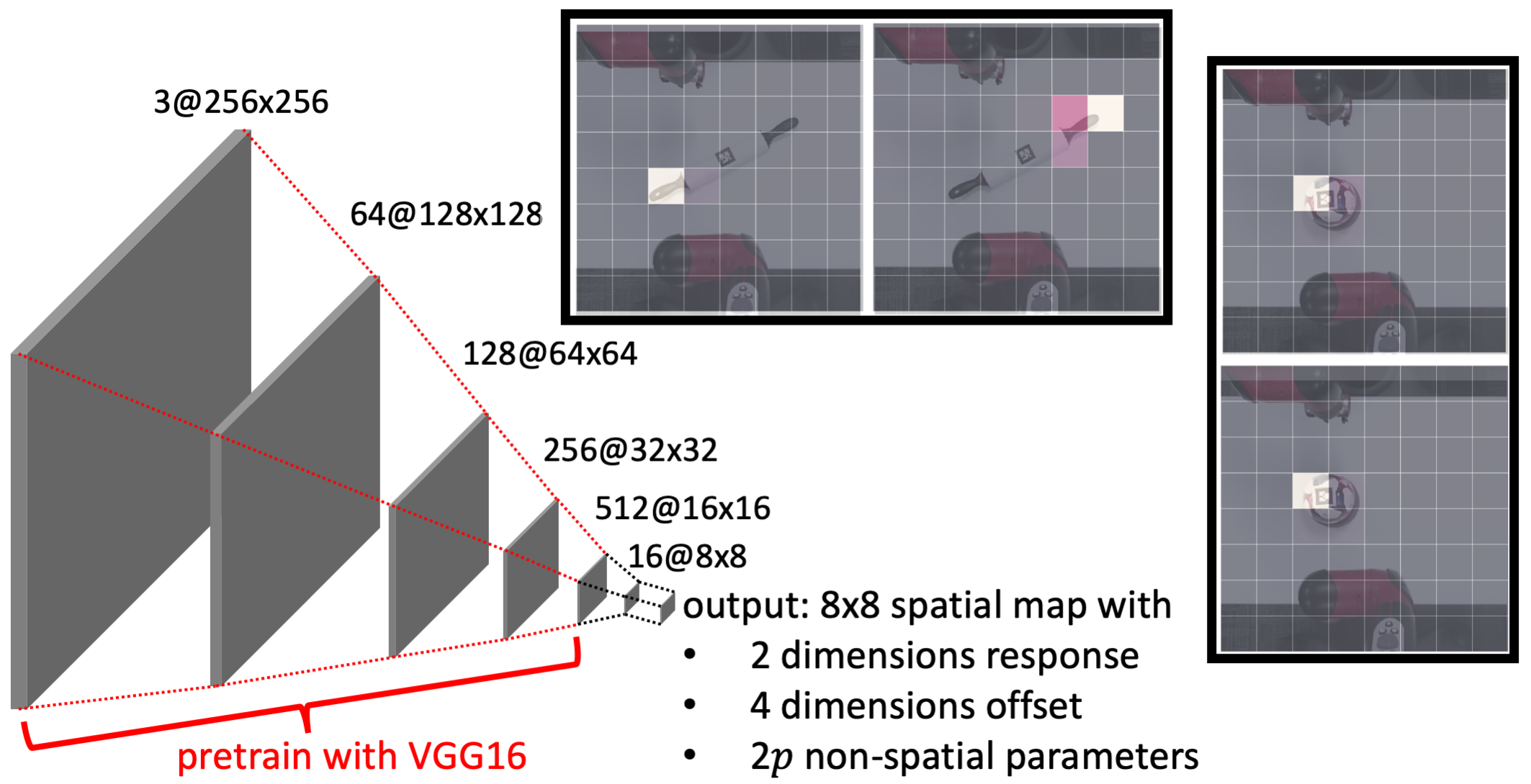}
\end{center}
\caption{To predict continuous arguments given an image input, we
  leverage the fact that all tasks require learning two spatial
  arguments for each arm: an (x, y) location along the table
  surface. To learn efficiently, we use a fully convolutional
  architecture~\cite{fullyconv}. We begin by passing the image through
  the first four layers of an ImageNet-pretrained VGG16~\cite{vgg};
  each layer is 2 convolutions followed by a max-pool. This gives us
  512 $8\times 8$ maps. The second-last layer convolves with 16
  $2\times 2$ filters with stride 2, giving 16 $8\times 8$
  maps. Finally, the last layer convolves with $(2+4+2p)$ filters with
  stride 1, where $p$ is the number of non-spatial parameters for the
  task. The first two dimensions give each arm's probability
  distribution (response map) over an $8\times 8$ discretization of
  the table surface locations, the next four dimensions give offsets
  on these locations, and the final $2p$ dimensions give values for
  each arm's remaining arguments (orientations, distances,
  etc.). \emph{Upper right:} Example response maps learned by this
  model for each arm, for the rolling pin and soccer ball tasks. The
  AR tags are only used in our experiments for automating the reward
  computation to measure success; our model is not given any object
  pose information.}
\label{fig:spatial}
\end{figure}

\begin{figure}
\begin{center}
\includegraphics[width=\linewidth]{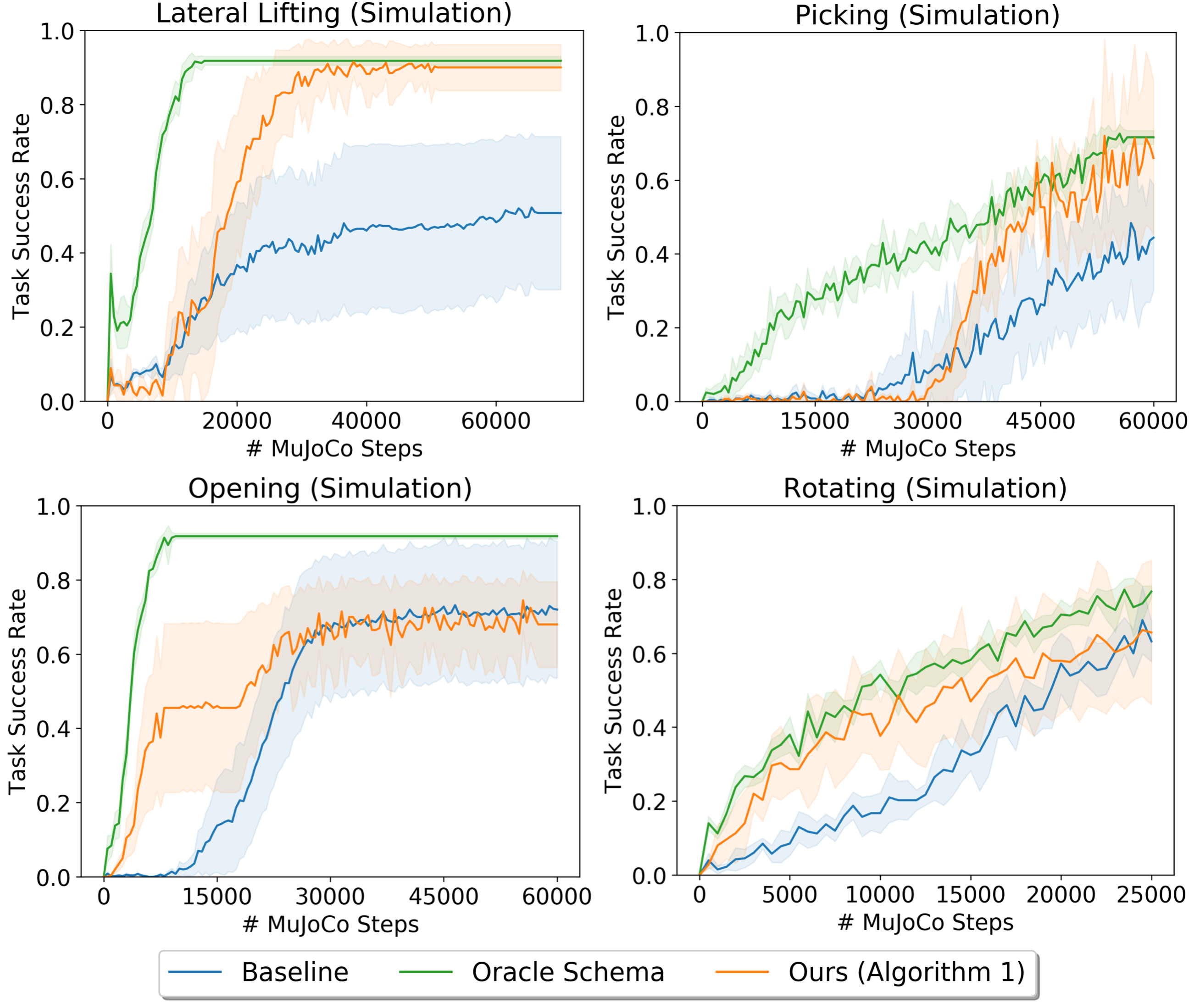}
\end{center}
\caption{Learning curves for simulated tasks, with training
  parallelized across 50 workers. Each curve depicts an average across
  5 random seeds. By using a policy architecture that leverages the
  state-independence of the optimal schema (orange), we are able to
  achieve significant gains in sample complexity across all four
  tasks, over the baseline architecture that predicts skills and
  arguments conditioned on the state (blue). If an oracle tells us the
  perfect schema, and we only need to learn the arguments for those
  skills, then of course, learning will be extremely sample-efficient
  (green).}
\label{fig:resultsschem}
\end{figure}

\begin{figure}[t]
  \vspace{0.7em}
\begin{center}
\includegraphics[width=\linewidth]{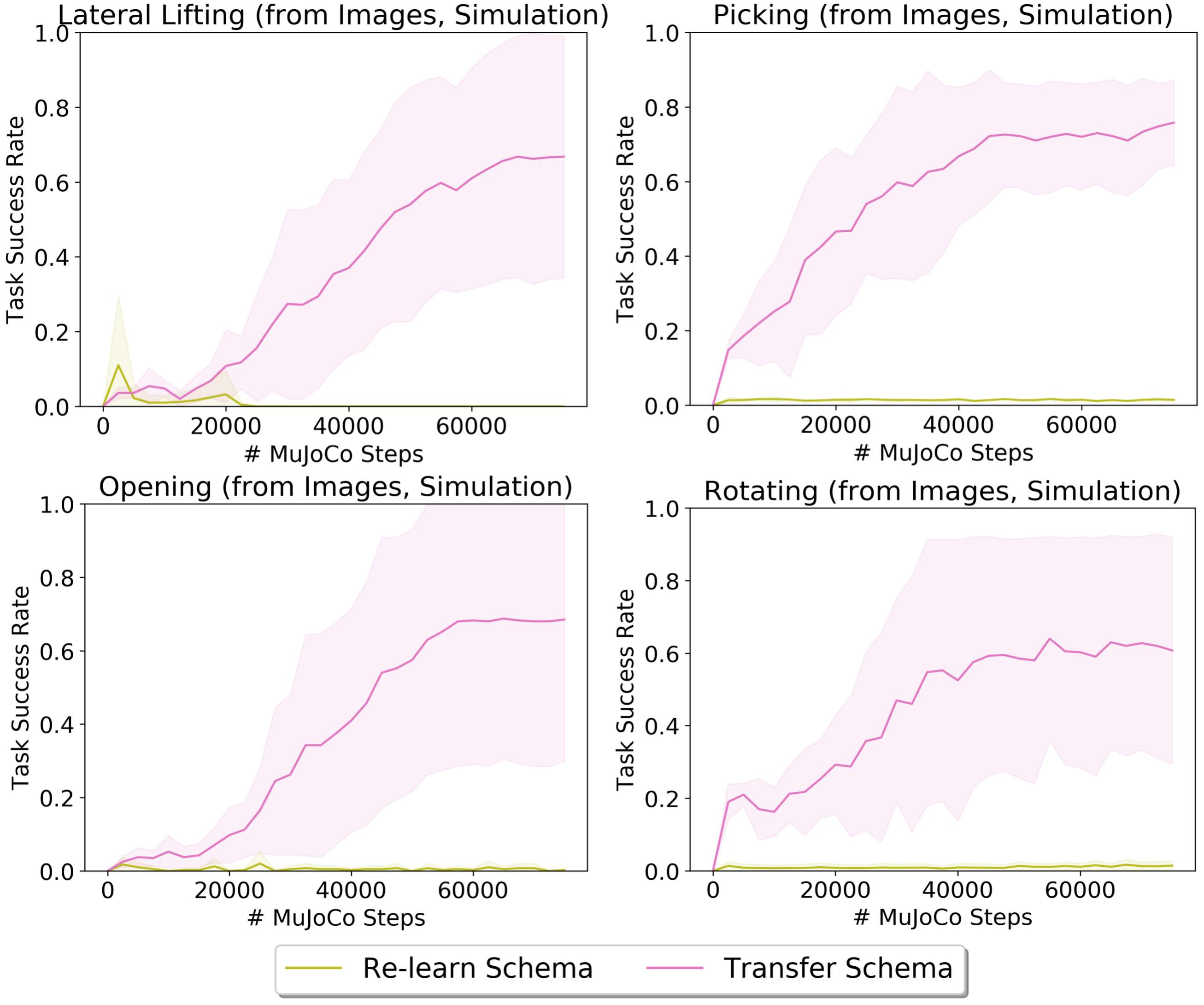}
\end{center}
\caption{Learning curves for training from images in simulation, with
  training parallelized across 50 workers. Each curve depicts an
  average across 5 random seeds. When learning visual policies,
  transferring schemas trained on low-dimensional states is
  crucial.}
\label{fig:resultsimagesim}
\end{figure}

\begin{figure}[t]
  \vspace{0.7em}
\begin{center}
\includegraphics[width=\linewidth]{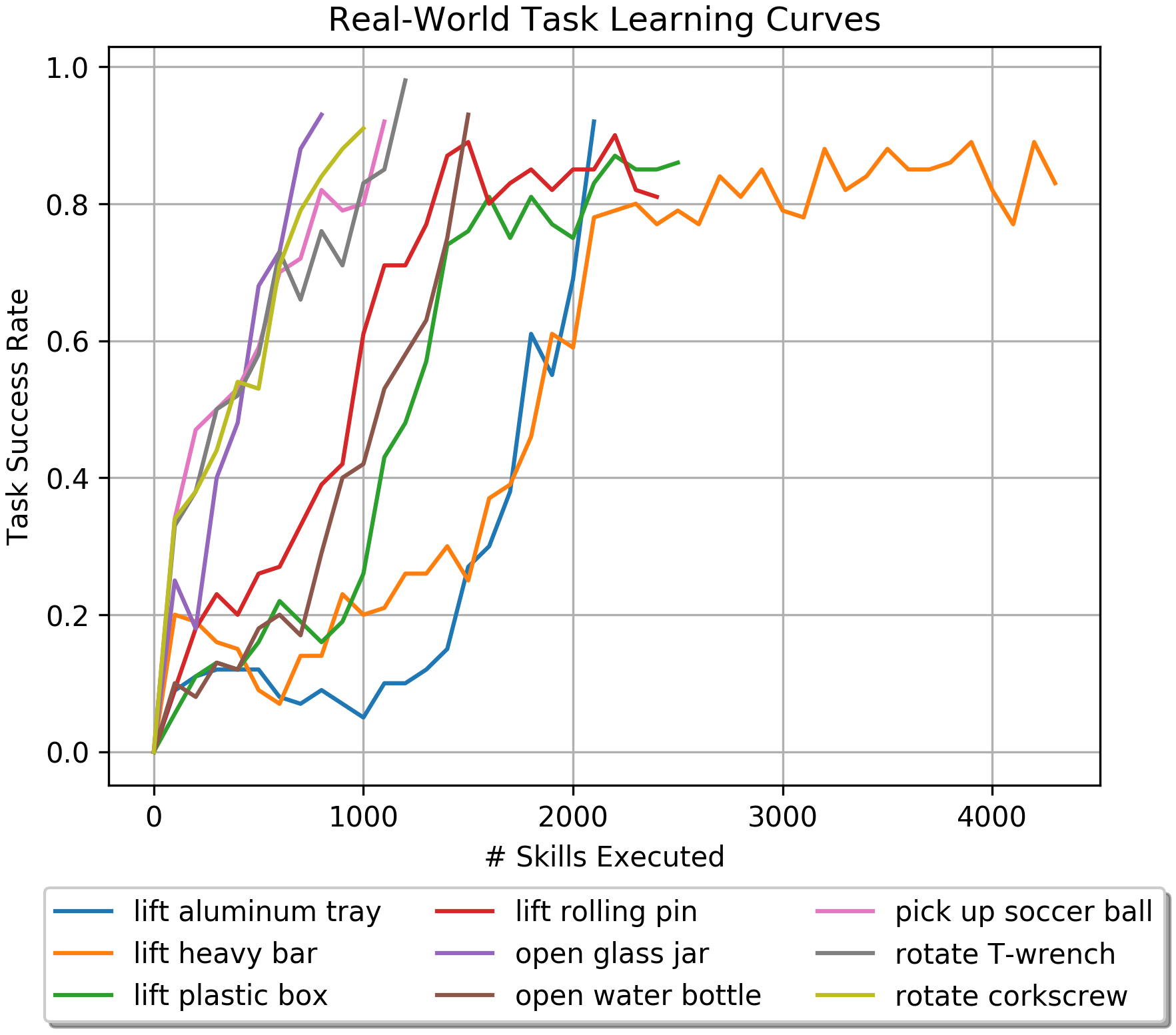}
\end{center}
\caption{Learning curves for the real-world tasks. We stop training
  when the policy reaches 90\% average success rate over the last 100
  episodes. It takes around 6-8 hours to execute 2000 skills, so most
  policies are learned in around 4-10 hours. By transferring schemas
  learned in simulation, we can train robots to solve sparse-reward
  bimanual manipulation tasks from raw camera images.}
\label{fig:resultsrealworld}
\end{figure}

\subsection{Real-World Sawyer Experimental Setup}
\vspace{1mm}
\noindent \textbf{Environment.} Our real-world setup also contains two
Sawyer robot arms with parallel-jaw grippers placed at opposing ends
of a table, facing each other. We task the robots with manipulating
nine common household objects that require two parallel-jaw grippers
to interact with. We consider the same four task families (lateral
lifting, picking, opening and rotating), but work with more diverse
objects (such as a rolling pin, soccer ball, glass jar, and
T-wrench), as detailed in \tabref{tab:schemas}. For each task family,
we use the schema discovered for that family in simulation, and only
learn the continuous parameterizations of the skills in the real
world. See \figref{fig:teaser} for pictures of some of our tasks.

\vspace{1mm}
\noindent \textbf{Skills.} The skills and parameters are the same as
in simulation (\tabref{tab:skills}), but the search spaces are less
constrained (\tabref{tab:parameters}) since we do not have access to
object poses.

\vspace{1mm}
\noindent \textbf{State and Policy Representation.} The state for
these real-world tasks is the $256\times 256$ RGB image obtained from
an overhead camera that faces directly down at the table. To predict
the continuous arguments, we use a fully convolutional spatial neural
network architecture~\cite{fullyconv}, as shown in
\figref{fig:spatial} along with example response maps.

\vspace{1mm}
\noindent \textbf{Training Details.} We use \ppo\ and mostly the same
hyper-parameters, with the following differences: learning rate
$0.005$, number of steps $500$, number of minibatches per update $10$,
number of optimization epochs $10$, and no parallelization. We control
the Sawyers using PyRobot~\cite{pyrobot}.

\subsection{Results in Simulation}

\figref{fig:resultsschem} shows that our policy architecture greatly
improves the sample efficiency of model-free reinforcement learning.
In all simulated environments, our method learns the optimal schema,
as shown in the last column of \tabref{tab:schemas}.  Much of the
difficulty in these tasks stems from sequencing the skills correctly,
and so our method, which more effectively shares experience across
task instantiations in its attempt to learn the task schema, performs
very well.

Before transferring the learned schemas to the real-world tasks, we
consider learning from rendered images in simulation, using the
architecture from \figref{fig:spatial} to process
them. \figref{fig:resultsimagesim} shows the impact of transferring
the schema versus re-learning it in this more realistic simulation
setting. We see that when learning visual policies, transferring the
schemas learned in the tasks with low-dimensional state spaces is
critical to efficient training. These results increase our confidence
that transferring the schema will enable efficient real-world training
with raw RGB images, as we show next.

\subsection{Results in Real World}

\figref{fig:resultsrealworld} shows our results on the nine real-world
tasks, with schemas transferred from the simulated tasks. We can see
that, despite the challenging nature of the problem (learning from raw
camera images, given sparse rewards), our system is able to learn to
manipulate most objects in around 4-10 hours of training. We believe
that our approach can be useful for sample-efficient learning in
problems other than manipulation as well; all one needs is to define
skills appropriate for the environment such that the optimal sequence
depends only on the task, not the (dynamic) state. The skills may
themselves be parameterized closed-loop policies.

\textbf{Please see the supplementary video for examples of learned
  behavior on the real-world tasks.}

%% file: exp-skills.tex
\begin{table}[t]
  \vspace{0.7em}
  \centering
  \resizebox{\columnwidth}{!}{
  \tabcolsep=0.08cm{
  \begin{tabular}{ccc}
    \toprule[1.5pt]
    \textbf{Skill} & \textbf{Allowed Task Families} & \textbf{Continuous Parameters}\\
    \midrule[2pt]
    top grasp & lateral lifting, picking, opening & (x, y) position, z-orientation\\
    \midrule
    side grasp & opening, rotating & (x, y) position, approach angle\\
    \midrule
    go-to pose & picking, rotating & (x, y) position, orientation\\
    \midrule
    lift & lateral lifting, picking & distance to lift\\
    \midrule
    twist & opening & none\\
    \midrule
    rotate & rotating & rotation axis, rotation radius\\
    \midrule
    no-op & all & none\\
    \bottomrule[1.5pt]
  \end{tabular}}}
  \caption{Skills, allowed task families, and skill parameters.}
  \label{tab:skills}
\end{table}

%% file: exp-skills-params.tex
\begin{table}[t]
  \centering
  \resizebox{\columnwidth}{!}{
  \tabcolsep=0.1cm{
  \begin{tabular}{ccp{3cm}p{3cm}}
    \toprule[1.5pt]
    \textbf{Parameter} & \textbf{Relevant Skills} & \textbf{Search Space (Sim)} & \textbf{Search Space (Real)}\\
    \midrule[2pt]
    (x, y) position & grasps, go-to pose & \specialcell{[-0.1, 0.1] x/y/z offset\\from object center} & location on table surface\\
    \midrule
    z-orientation & top grasp & $[0, 2\pi]$ & $[0, 2\pi]$ \\
    \midrule
    approach angle & side grasp & $[-\frac{\pi}{2}, \frac{\pi}{2}]$ & $[-\frac{\pi}{2}, \frac{\pi}{2}]$\\
    \midrule
    orientation & go-to pose & \specialcell{$[0, 2\pi]$ r/p/y Euler angles\\converted to quat} & \specialcell{$[0, 2\pi]$ r/p/y Euler angles\\converted to quat}\\
    \midrule
    distance to lift & lift & $[0, 0.5]$ & $[0, 0.5]$\\
    \midrule
    rotation axis & rotate & \specialcell{[-0.1, 0.1] x/y offset\\from object center} & location on table surface\\
    \midrule
    rotation radius & rotate & $[0, 0.2]$ & $[0, 0.2]$\\
    \bottomrule[1.5pt]
  \end{tabular}}}
  \caption{Parameter search spaces for simulation and real world. In simulation, unlike the real world, we have access to object poses, so we can constrain some search spaces for efficiency. Positions, distances, and lengths are in meters. \emph{Rotation axis} is always vertical.}
  \label{tab:parameters}
\end{table}

%% file: futurework.tex
\section{Future Work}
\label{sec:futurework}
In this work, we have studied how to leverage state-independent
sequences of skills to greatly improve the sample efficiency of
model-free reinforcement learning. Furthermore, we have shown
experimentally that transferring sequences of skills learned in
simulation to real-world tasks enables us to solve sparse-reward
problems from images very efficiently, making it feasible to train
real robots to perform complex skills such as bimanual manipulation.

An important avenue for future work is to relax the assumption that
the optimal schema is open-loop. For instance, one could imagine
predicting the schema via a recurrent mechanism, so that the decision
on what skill to use at time $t+1$ is conditioned on the skill used at
time $t$.  Another interesting future direction is to study
alternative approaches to training the state-independent schema
predictor.
Finally, we hope to investigate the idea of inferring the schemas from
human demonstrations of a task, such as those in videos.